\documentclass{article}

\usepackage{PRIMEarxiv}

\usepackage[utf8]{inputenc}
\usepackage[T1]{fontenc}
\usepackage{hyperref}
\usepackage{url}
\usepackage{booktabs}
\usepackage{placeins}
\usepackage{amsfonts}
\usepackage{amsmath}
\usepackage{amssymb}
\usepackage{nicefrac}
\usepackage{microtype}
\usepackage{fancyhdr}
\usepackage{graphicx}
\usepackage{algorithm}
\usepackage{algpseudocode}
\graphicspath{{media/}}
\hypersetup{hypertexnames=false}

\title{Bridging Diffusion Pruning and Step Distillation with Teacher-Aligned Repair}

\author{
  Jincheng Ying,
  Li Wenlin \\
  School of Big Data and Artificial Intelligence \\
  Guangdong University of Finance and Economics \\
  Guangzhou, China \\
  \texttt{jc\_ying@student.gdufe.edu.cn},\\
  \texttt{wenlin@gdufe.edu.cn} \\          % 请确认并替换为真实邮箱
  \And
  Minghui Xu \\
  School of Computer Science \\
  Shandong University \\
  Qingdao, China \\
  \texttt{mhxu@sdu.edu.cn} \\
  \And
  Yinhao Xiao\thanks{Corresponding author.} \\
  GDUFE-UCM Joint Institute \\
  Guangdong University of Finance and Economics \\
  Guangzhou, China \\
  \texttt{20191081@gdufe.edu.cn} \\
}

\begin{document}
\maketitle

\begin{abstract}
Diffusion models generate high-quality images, but their inference cost comes from two sources: large denoising networks and repeated denoising steps. Existing compression pipelines usually attack these costs separately. Pruning reduces the network, but most pruning methods still rely on a long post-pruning retraining stage to recover a many-step sampler. Step distillation reduces the number of denoising steps, but it usually assumes a student that can already follow the teacher well enough to receive useful distillation gradients. This paper asks whether post-pruning retraining can be replaced by step distillation. We find that the direct replacement fails: after pruning an EDM2-XS teacher, starting SiDA from the pruned checkpoint produces unusable samples. We introduce a short teacher-alignment repair stage as a bridge between pruning and step distillation. The bridge matches the pruned generator to the teacher on noisy real-image latents, then hands the repaired checkpoint to one-step distillation. On ImageNet-512, the original EDM2-XS baseline uses 124.713M parameters and 63 network evaluations, reaching an FID of 3.53. With a suitable distillation objective, our 20\% pruned one-step generator uses 98.826M parameters and one network evaluation, reaching an FID of 3.12. With 30\% pruning, the model uses 88.029M parameters and one network evaluation, with an FID of 4.26.
\end{abstract}


\section{Introduction}

Diffusion models have become a central tool for high-quality image synthesis, including unconditional generation, class-conditional generation, text-to-image generation, editing, and super-resolution~\cite{ho2020ddpm,dhariwal2021diffusion,rombach2022ldm,karras2024edm2}. Their cost has two sources. The denoising network is large, and sampling evaluates it repeatedly. On ImageNet-512, the EDM2-XS sampler in our setup uses 63 network evaluations per image.

Most efficiency work reduces one of these two costs. Pruning removes weights, channels, blocks, or heads from a pretrained diffusion model~\cite{fang2023diffpruning,castells2024ldpruner,ramesh2025efficient}. Step distillation reduces the number of denoising evaluations, sometimes to one or a few steps~\cite{salimans2022progressive,song2023consistency,yin2024dmd,zhou2024sida}. These directions are complementary, but they are usually not connected. A pruned diffusion model is commonly recovered by retraining or fine-tuning the pruned network as a many-step sampler. Step distillation, in contrast, is normally applied to a student that is already compatible with the teacher's denoising field.

This raises a direct question: can the expensive post-pruning retraining stage be replaced by step distillation? If so, pruning would reduce the network size and distillation would reduce the sampling trajectory in the same recovery process. The recovered model would not be a smaller many-step denoiser. It would be a compact one-step generator.

The direct replacement does not work in our experiments. We prune an EDM2-XS teacher and start SiDA one-step distillation~\cite{zhou2024sida} from the pruned checkpoint. The model fails to produce usable samples, as shown in Table~\ref{tab:bridge-motivation}. This failure should not be read as a limitation of SiDA alone. After structured pruning, the student can be far from the teacher on the noisy states where a teacher-based distillation loss is evaluated. The distillation method then receives a damaged initialization rather than a compact student ready to learn.

\begin{table}[t]
  \caption{Replacing post-pruning retraining with step distillation is not a direct plug-in. The pruned checkpoint fails under direct one-step distillation, while the proposed repair bridge makes the same recovery path usable.}
  \centering
  \resizebox{\textwidth}{!}{
  \begin{tabular}{lcccc}
    \toprule
    Model & Repair & NFE & \#Params & FID ($\downarrow$) \\
    \midrule
    EDM2-XS teacher~\cite{karras2024edm2} & -- & 63 & 124.713M & 3.53 \\
    Pruned + direct SiDA & No & 1 & 88.029M & 112.56 (failed) \\
    Ours (20\% pruned) & Yes & 1 & 98.826M & 3.12 \\
    Ours (30\% pruned) & Yes & 1 & 88.029M & 4.26 \\
    \bottomrule
  \end{tabular}
  }
  \label{tab:bridge-motivation}
\end{table}

We introduce a bridge between pruning and step distillation: a short teacher-alignment repair stage. Given a pruned generator, repair matches its output to the teacher on noisy real-image latents in the high-noise region used by one-step generation. It is not a full retraining run. It only moves the pruned generator into a region where a teacher-based step-distillation loss has useful gradients. The repaired checkpoint is then used to initialize one-step distillation.

With a suitable distillation objective, the repaired pruned model can be both smaller and one-step. On EDM2-XS/ImageNet-512, the 20\% pruned generator improves FID from 3.53 to 3.12 while reducing NFE from 63 to 1. A 30\% pruned generator keeps one-step generation with 88.029M parameters and an FID of 4.26.

Our main contributions are as follows:
\begin{itemize}
  \item We identify post-pruning retraining as a bottleneck for diffusion compression and show that a direct replacement by one-step distillation fails after structured pruning.
  \item We propose a teacher-alignment repair bridge that connects pruned diffusion architectures to teacher-based step distillation without a full retraining run.
  \item On EDM2-XS/ImageNet-512, the repaired 20\% pruned generator improves FID from 3.53 to 3.12 while reducing NFE from 63 to 1; the 30\% pruned generator keeps one-step generation with 88.029M parameters.
\end{itemize}

\section{Related Work}

\subsection{Pruning Diffusion Models}

Pruning removes parameters, channels, blocks, or heads from a pretrained model. The usual pipeline then retrains or fine-tunes the remaining network so that it can compensate for the removed structure. This pattern goes back to early neural network pruning, where weights and connections are removed and the remaining network is retrained~\cite{han2015learning}. Diffusion pruning inherits the same recovery problem, but the cost is higher: the pruned model must still preserve a denoising field over many noise levels.

Diff-Pruning uses a Taylor expansion over diffusion timesteps to identify less important structure and recover a lightweight diffusion model without extensive retraining~\cite{fang2023diffpruning}. LD-Pruner measures the effect of pruning in latent space and uses this signal to prune latent diffusion models across text, image, and audio generation tasks~\cite{castells2024ldpruner}. Recent work on Stable Diffusion studies post-training pruning and reports that quality can break abruptly when pruning passes certain sparsity levels~\cite{ramesh2025efficient}. Learnable-sparsity pruning goes further toward low-cost recovery by learning differentiable structural masks and avoiding a separate retraining phase~\cite{zhang2024effortless}. These methods reduce the cost of pruning recovery, but their output is still a pruned diffusion sampler. Sampling remains iterative.

Our goal is different from recovering a smaller many-step denoiser. We use pruning as the architecture reduction step and step distillation as the recovery step, so the output of the pipeline is a compact one-step generator. This changes the role of post-pruning training. If pruning is followed by a long retraining run, the compression pipeline still pays to recover a many-step denoising process before any sampling-step reduction is obtained. We instead use a short teacher-alignment repair and then train the compact model with one-step distillation.

\subsection{Diffusion Distillation}

Diffusion distillation reduces the number of sampling steps. Progressive distillation trains a student sampler to replace two teacher steps with one, and repeats this halving procedure until only a few steps remain~\cite{salimans2022progressive}. Consistency models learn mappings that are consistent across noise levels and support one-step or few-step generation~\cite{song2023consistency}. Distribution Matching Distillation trains a one-step generator by matching the distribution of a multi-step diffusion model with a score-difference objective~\cite{yin2024dmd}. SiD and SiDA use score identity losses; SiDA adds real images and adversarial training to improve one-step distillation from EDM and EDM2 teachers~\cite{zhou2024sida}.

These methods mainly address sampling speed. They usually assume that the student has enough capacity and starts from an initialization suitable for distillation. Pruning changes both conditions: the student is smaller, and its local denoising field can be misaligned with the teacher. Direct one-step distillation from the pruned checkpoint fails in our experiments. The repair stage is introduced to make the pruned student a usable starting point for the downstream distillation loss.

\subsection{Pruning with Distillation}

Pruning and knowledge distillation are often combined: pruning reduces capacity, and distillation helps the smaller model imitate a larger teacher. BK-SDM removes residual and attention blocks from Stable Diffusion and uses feature-level knowledge distillation to recover compact text-to-image models~\cite{kim2023bksdm}. It does not use step distillation, so the recovered model is still used as a multi-step diffusion sampler. SnapFusion combines a compact U-Net, decoder optimization, and improved step distillation to run text-to-image generation on mobile devices~\cite{li2023snapfusion}. Its design is an effective system recipe for a specific mobile text-to-image pipeline, but the pruning and step-distillation interface is not formulated as a general recovery framework for pruned diffusion architectures.

Our setting combines the two directions at a different interface. We start from the post-pruning recovery bottleneck and ask whether the retraining stage can be replaced by step distillation. Direct replacement does not work, so we insert a short repair stage before distillation. Repair is not intended to solve generation by itself. It reduces local teacher-student denoising mismatch so that a pruned architecture can enter a teacher-based step-distillation objective. This gives a general way to connect pruning with one-step or few-step distillation, rather than treating pruning recovery and sampling-step reduction as separate recipes.

\section{Method}
\label{sec:method}

Our goal is to replace the usual post-pruning retraining stage with a distillation stage. The recovered model should be a compact one-step generator, not a smaller many-step denoiser. The argument is independent of the exact distillation loss. We assume only that the downstream distillation method trains a compact generator $G_\theta$ against a fixed teacher $T$ through some loss $\mathcal{D}(G_\theta;T,\mathcal{A})$.

The full procedure has four steps. We first prune the EDM2 teacher into a smaller generator architecture. We then test the direct path: using this pruned model as the distillation initialization. This path fails because the pruned model is not close enough to the teacher's denoising behavior. We then repair the pruned model by aligning it to the teacher on noisy real latents. Finally, we run one-step distillation from the repaired checkpoint. In our implementation the distillation method is SiDA, but the repair stage is defined before SiDA and can be paired with other teacher-based distillation losses.

\subsection{From Post-Pruning Retraining to Distillation}

Let $T$ denote the pretrained EDM2 teacher. Standard pruning would build a smaller denoiser $T_{\mathcal{S}}$ and then retrain or fine-tune $T_{\mathcal{S}}$ as a diffusion model. The recovery objective remains tied to denoising:
\begin{equation}
  T_{\mathcal{S}} \leftarrow
  \arg\min_{\theta}
  \mathbb{E}_{x,y,\sigma,\epsilon}
  \left[
  \left\|
  T_{\theta}(x+\sigma\epsilon,\sigma,y) - x
  \right\|_2^2
  \right].
\end{equation}
Sampling still needs a reverse diffusion trajectory after this recovery.

We instead want the recovery model to be a one-step generator $G_\theta$. Given noise $\eta$ and label $y$, the generator returns a latent sample in one call:
\begin{equation}
  \hat{z}=G_\theta(\eta,\sigma_{\mathrm{init}},y).
\end{equation}
Let $G_0$ be the pruned model after weight copying. A direct replacement uses $G_0$ as the initialization for a distillation loss:
\begin{equation}
  G_\theta \leftarrow \arg\min_{\theta:\theta_0=G_0}
  \mathcal{D}(G_\theta;T,\mathcal{A}).
\end{equation}
This is only a good replacement for retraining if $G_0$ lies in the basin where the distillation loss gives useful gradients. After structured pruning, that condition can fail. The pruned model has the correct parameter count, but its denoising outputs can be far from the teacher on noisy states that the distillation method later uses.

We formalize this mismatch with a local teacher-alignment error
\begin{equation}
  \varepsilon(G;\mu)
  =
  \mathbb{E}_{(\tilde{z},\sigma,y)\sim \mu}
  \left[
  \left\|G(\tilde{z},\sigma,y)-T(\tilde{z},\sigma,y)\right\|_2^2
  \right],
\end{equation}
where $\mu$ is a distribution over noisy latent states. If $\varepsilon(G_0;\mu)$ is large, the student and teacher define different local denoising fields. A score-based, adversarial, consistency, or regression-based distillation loss may all suffer from this bad initialization, because each uses teacher information to shape the generator. The role of repair is to find a nearby initialization $G_{\mathrm{align}}$ with lower $\varepsilon(G_{\mathrm{align}};\mu)$ before optimizing $\mathcal{D}$.

This failure motivates the teacher-alignment repair in Section~\ref{sec:repair}. The repair stage is not the original post-pruning retraining. It does not train a many-step sampler. It only reduces $\varepsilon(G;\mu)$ on a small set of noisy real latents, so that one-step distillation starts from a model with a meaningful denoising field.

\subsection{Compact Generator Parameterization}

Let $T$ be the pretrained EDM2 teacher and let $G$ be the one-step generator architecture used by SiDA. The compact generator keeps the same input and output interface as the teacher. It takes a noisy latent, a noise level, and a class label, and it returns a denoised latent. Only the internal architecture is smaller.

We use two structured pruning operations. The first changes channel widths at each U-Net resolution:
\begin{equation}
  (c_1,c_2,c_3,c_4),
\end{equation}
where each $c_i$ is the width at one resolution. The second removes selected branches from U-Net blocks. This is not a binary mask. When a branch is removed, the corresponding convolution, embedding projection, attention projection, and gain parameters are not instantiated in the compact model.

For a block $b$, branch removal changes the block computation from
\begin{equation}
  h_{b+1} = \mathrm{MPsum}\left(h_b, R_b(h_b, e)\right) + A_b(h_b)
\end{equation}
to the surviving skip/resample path. Here $R_b$ denotes the residual branch, $A_b$ denotes self-attention, and $e$ is the class/noise embedding. In implementation, both $R_b$ and $A_b$ are disabled for selected full-block pruning candidates. This gives real parameter reduction and avoids carrying zeroed tensors through inference.

The channel-pruned variant uses explicit channel lists such as $(104,208,384,512)$ or $(104,208,312,416)$. When widths differ from the teacher, teacher tensors are sliced into the smaller tensors. Decoder blocks need special handling because their input is a concatenation of the main path and the skip path. We slice these two parts separately and then concatenate the selected indices. This keeps the U-Net skip layout valid after pruning.

\subsection{Teacher-Aware Block Sensitivity}

Width pruning alone can remove many parameters, but it does not tell us which denoising computations are safe to remove before distillation. The recovery problem gives a natural importance function. If removing a block makes the pruned model's denoising output move far away from the teacher, distillation starts from a bad point. If the output barely changes, the block is a better pruning candidate.

We first draw a fixed calibration set of real images, encode them to latents, sample noise at a chosen noise level $\sigma^\star$, and compute teacher targets
\begin{equation}
  z_i = E(x_i), \qquad
  \tilde{z}_i = z_i + \sigma^\star \epsilon_i, \qquad
  t_i = T(\tilde{z}_i, \sigma^\star, y_i),
\end{equation}
where $E$ is the VAE encoder. In our ImageNet experiments, $\sigma^\star=2.5$.

The calibration set defines an empirical distribution $\hat{\mu}_{\sigma^\star}$ over noisy states at a fixed noise level. For a compact student $G$, the empirical teacher-alignment error is
\begin{equation}
  \hat{\varepsilon}_{\sigma^\star}(G)
  =
  \frac{1}{N}\sum_{i=1}^{N}
  \left\|
  G(\tilde{z}_i, \sigma^\star, y_i) - t_i
  \right\|_2^2 .
\end{equation}
Let $G^{-b}$ be the same model with candidate block $b$ temporarily removed. The importance of $b$ is the increase in teacher-alignment loss:
\begin{equation}
  I(b)
  =
  \hat{\varepsilon}_{\sigma^\star}(G^{-b})
  -
  \hat{\varepsilon}_{\sigma^\star}(G).
\end{equation}
Expanding the difference shows what the score measures:
\begin{equation}
  I(b)
  =
  \frac{1}{N}\sum_{i=1}^{N}
  \left(
  \left\|G^{-b}(\tilde{z}_i,\sigma^\star,y_i)-t_i\right\|_2^2
  -
  \left\|G(\tilde{z}_i,\sigma^\star,y_i)-t_i\right\|_2^2
  \right).
\end{equation}
Small $I(b)$ means that removing block $b$ does little damage to the local teacher-student denoising relation used by the repair stage. We rank candidate blocks by $I(b)$ and remove low-importance blocks until the parameter budget is reached. The exact block names and channel widths are implementation parameters and are reported with the experimental setup.

\subsection{Channel Importance for Width-Pruned Initialization}

We also use a channel-pruned initialization in some variants. For each trainable tensor in the teacher, we compute a Taylor-style channel importance score from EDM denoising loss:
\begin{equation}
  s_j \leftarrow s_j + \sum_{k \in \Omega_j} \left|w_k \frac{\partial \mathcal{L}_{\mathrm{EDM}}}{\partial w_k}\right|,
\end{equation}
where $\Omega_j$ is the set of weights associated with channel $j$. Scores are accumulated over calibration mini-batches and normalized per dimension. A fraction of prefix channels is protected, and the remaining channels are selected by top-$k$ importance. Prefix protection keeps the EDM2 channel layout and attention head structure stable. This width-pruned path is useful as a comparison and as a source model for later block pruning, but our best result comes from block sensitivity pruning directly from the teacher.

\subsection{Noisy-Real Teacher Alignment}
\label{sec:repair}

After pruning, $G_{\mathcal{S}}$ has the desired size, but direct distillation does not recover image quality. The pruned model has lost part of the teacher's local denoising field. A downstream distillation loss then starts from samples and noisy states where the compact model and the teacher disagree. We therefore align $G_{\mathcal{S}}$ to the teacher before using it as a one-step generator.

The alignment is deliberately short. It is not a new diffusion training run, and it does not try to make the pruned model a full many-step sampler. It only asks the pruned model to agree with the teacher at a small set of noisy latent states. Let $z=E(x)$ be a real-image latent. We sample noise at a fixed repair level $\sigma_r$ and match the teacher output:
\begin{equation}
  \tilde{z} = z + \sigma_r \epsilon,
\end{equation}
\begin{equation}
  \mathcal{L}_{\mathrm{repair}}(G_{\mathcal{S}})
  =
  \mathbb{E}_{x,y,\epsilon}
  \left[
  \left\|
  G_{\mathcal{S}}(\tilde{z}, \sigma_r, y)
  -
  T(\tilde{z}, \sigma_r, y)
  \right\|_2^2
  \right].
\end{equation}
This objective is exactly the local error $\varepsilon(G_{\mathcal{S}};\mu)$ for the distribution $\mu$ induced by real latents and the fixed noise level $\sigma_r$. If $\theta_{\mathcal{S}}$ are the parameters after pruning, a short alignment run gives
\begin{equation}
  \theta_{\mathrm{align}}
  =
  \theta_{\mathcal{S}}
  -
  \tau
  \nabla_{\theta}
  \mathcal{L}_{\mathrm{repair}}(G_{\theta})
  \big|_{\theta=\theta_{\mathcal{S}}}
  \quad
  \text{for a small number of updates}.
\end{equation}
The notation only indicates the direction of the update; in practice we use Adam for a short calibration run.

The choice of $\sigma_r$ follows the role of the repaired model. The one-step generator starts from high-noise latents with scale $\sigma_{\mathrm{init}}$. If the pruned model is wrong in this high-noise region, the first generated latent is already poor and later distillation has little useful signal. We therefore align at the largest relevant noise level:
\begin{equation}
  \sigma_r
  =
  \max\{\sigma_{\mathrm{init}}, \sigma_1,\ldots,\sigma_K\},
\end{equation}
where $\{\sigma_k\}$ are optional diagnostic noise bins used during repair. In our runs, $\sigma_{\mathrm{init}}=2.5$ and the bins are below it, so $\sigma_r=2.5$. This choice makes repair focus on the region where one-step generation begins. It also avoids turning repair into a full multi-noise retraining procedure.

We do not use the generator prior during this stage. The data are real latents, and the target is the teacher denoising prediction at the same noisy latent.

This stage is the bridge between pruning and distillation. Unlike post-pruning retraining, it does not try to recover a complete many-step sampler. Its target is narrower: reduce the local teacher-student mismatch on the noisy states that initialize one-step generation. Once this mismatch is reduced, the repaired checkpoint is used as the starting point for downstream distillation rather than as the final training objective.

\subsection{One-Step Distillation after Alignment}

The aligned compact generator $G_{\theta_{\mathrm{align}}}$ initializes the downstream distillation method:
\begin{equation}
  \theta_0 \leftarrow \theta_{\mathrm{align}},
  \qquad
  \theta^\star
  =
  \arg\min_{\theta:\theta_0=\theta_{\mathrm{align}}}
  \mathcal{D}(G_\theta;T,\mathcal{A}).
\end{equation}
The repair stage is useful when $\mathcal{D}$ is locally stable near a teacher-aligned initialization. Suppose the first-order change of the distillation loss around the initialization is controlled by the local teacher mismatch:
\begin{equation}
  \left\|\nabla_\theta \mathcal{D}(G_\theta;T,\mathcal{A})\right\|
  \leq
  C_1 + C_2 \sqrt{\varepsilon(G_\theta;\mu)}
\end{equation}
for the noisy-state distribution $\mu$ used by the distillation method. This is a mild way to state the assumption used in practice: when the student and teacher disagree strongly on the states seen by distillation, the update direction is noisy or unhelpful; when they agree locally, the distillation method can refine the generator. Teacher alignment reduces the second term before distillation begins.

In our implementation, $\mathcal{D}$ is SiDA. The true score network remains the frozen EDM2 teacher $T$. The fake-score network $F_\psi$ is initialized from the teacher and trained with the usual fake-score denoising loss. When adversarial training is enabled, the encoder of $F_\psi$ also outputs discriminator logits for real and generated latents.

For generator updates, the SiD identity term compares the teacher and fake-score predictions on noisy generated latents. In simplified notation, for generated latent $\hat{z}=G_\theta(\eta,\sigma_{\mathrm{init}},y)$ and noisy latent $\hat{z}_\sigma=\hat{z}+\sigma\epsilon$, the generator loss uses
\begin{equation}
  \left(T(\hat{z}_\sigma,\sigma,y)-F_\psi(\hat{z}_\sigma,\sigma,y)\right)
  \left[
  T(\hat{z}_\sigma,\sigma,y)-\hat{z}
  -
  \alpha
  \left(T(\hat{z}_\sigma,\sigma,y)-F_\psi(\hat{z}_\sigma,\sigma,y)\right)
  \right],
\end{equation}
with the normalization used in SiDA. The adversarial generator loss is added after an initial warm-up. A different teacher-based one-step distillation loss could replace this SiDA objective. The repair stage would still serve the same purpose: reduce local teacher-student mismatch before the compact generator is optimized as a one-step model.

\begin{algorithm}[t]
  \caption{Teacher-aligned pruning-to-distillation}
  \label{alg:teacher-aligned-pruning}
  \begin{algorithmic}[1]
    \Require pretrained teacher $T$; candidate block set $\mathcal{B}$; target budget $P$
    \Require calibration images $\mathcal{C}$; calibration noise $\sigma^\star$; repair noise candidates $\{\sigma_k\}_{k=1}^{K}$
    \Require one-step initialization noise $\sigma_{\mathrm{init}}$; distillation loss $\mathcal{D}(\cdot;T,\mathcal{A})$
    \Ensure compact one-step generator $G_{\theta^\star}$
    \State Encode calibration images to latents $z_i=E(x_i)$ and form noisy states $\tilde{z}_i=z_i+\sigma^\star\epsilon_i$.
    \State Compute teacher targets $t_i=T(\tilde{z}_i,\sigma^\star,y_i)$.
    \ForAll{$b\in\mathcal{B}$}
      \State Temporarily remove block or branch $b$ from the compact candidate to obtain $G^{-b}$.
      \State Estimate
      \[
        I(b)=
        \frac{1}{N}\sum_i
        \left(
        \|G^{-b}(\tilde{z}_i,\sigma^\star,y_i)-t_i\|_2^2
        -
        \|G(\tilde{z}_i,\sigma^\star,y_i)-t_i\|_2^2
        \right).
      \]
    \EndFor
    \State Select low-importance blocks until the compact architecture satisfies budget $P$.
    \State Build $G_{\theta_{\mathcal{S}}}$ by instantiating the compact architecture and copying or slicing teacher weights.
    \State Set $\sigma_r=\max\{\sigma_{\mathrm{init}},\sigma_1,\ldots,\sigma_K\}$.
    \For{$m=1,\ldots,M_{\mathrm{repair}}$}
      \State Sample $(x,y)$ from calibration data, encode $z=E(x)$, and draw $\epsilon\sim\mathcal{N}(0,I)$.
      \State Form $\tilde{z}=z+\sigma_r\epsilon$.
      \State Update $\theta$ using
      \[
        \nabla_{\theta}
        \left\|
        G_{\theta}(\tilde{z},\sigma_r,y)
        -
        T(\tilde{z},\sigma_r,y)
        \right\|_2^2 .
      \]
    \EndFor
    \State Let $\theta_{\mathrm{align}}$ be the repaired parameters.
    \State Initialize downstream one-step distillation with $\theta_0=\theta_{\mathrm{align}}$.
    \State Optimize $\theta^\star=\arg\min_{\theta:\theta_0=\theta_{\mathrm{align}}}\mathcal{D}(G_{\theta};T,\mathcal{A})$.
    \State \Return $G_{\theta^\star}$
  \end{algorithmic}
\end{algorithm}

\section{Experiments}

We evaluate whether post-pruning retraining can be replaced by teacher-aligned one-step distillation. The direct path initializes SiDA from the pruned checkpoint without repair. This path fails qualitatively: early samples collapse and the model does not produce usable images. With teacher alignment, the same compact architecture becomes trainable under SiDA. On ImageNet-512, the repaired compact generator reaches an FID of 3.12 with one network evaluation, improving over the original EDM2-XS baseline while using fewer parameters.

\subsection{Datasets}

The main experiments use ImageNet-512 with class labels. Images are encoded into the EDM2 latent space with the same VAE interface used by the teacher and by SiDA. Calibration batches for pruning and repair are sampled from real training images. We report FID, parameter count, and number of function evaluations. When runtime is reported, it is measured on a single RTX 3090.

\subsection{Pruning and Distillation Setup}

We use EDM2-XS as the teacher and construct compact generators with the teacher-aware pruning rule described in Section~\ref{sec:method}. The more compressed model in Table~\ref{tab:benchmark} removes low-importance residual-attention branches from low-resolution U-Net stages, while keeping the input and output interface unchanged. After pruning, we compare two recovery paths. The first path directly starts SiDA from the pruned checkpoint. The second path first repairs the pruned generator against the teacher on noisy real latents, then starts SiDA from the repaired checkpoint.

\subsection{Implementation Details}

For ImageNet-512, the downstream distillation stage uses SiDA with the frozen EDM2 teacher as the true score model and a fake-score network initialized from the teacher. Sampling from the final compact generator uses one generator evaluation in latent space followed by VAE decoding. The run-level pruning, repair, optimizer, and CIFAR-10 Diff-Instruct settings are listed in Appendix~\ref{app:reproducibility}.

\subsection{Ablation Study}

Table~\ref{tab:ablations} isolates the main design choices in the pruning-to-distillation pipeline. All rows use EDM2-XS as the teacher and SiDA as the downstream one-step distillation loss. Rather than listing training runs, the table changes one component at a time when possible: the repair bridge, the pruning form, the block-selection score, and the repair noise used for alignment. Parameter counts and runtime are reported in the benchmark table; the ablation table focuses on what each component contributes.

\begin{table}[t]
  \caption{Ablation study on ImageNet-512. The table reports FID for design variants of the pruning-to-distillation pipeline.}
  \centering
  \small
  \begin{tabular}{p{0.22\textwidth}p{0.60\textwidth}c}
    \toprule
    Component & Variant & FID ($\downarrow$) \\
    \midrule
    Full pipeline & Teacher-aware block pruning, high-noise repair, and SiDA distillation & \textbf{3.16} \\
    Repair bridge & Direct SiDA from the pruned checkpoint, without teacher-alignment repair & 112.56 \\
    Repair bridge & Use only a limited teacher-alignment repair before SiDA & 17.23 \\
    Pruning form & Replace block pruning with channel pruning while preserving attention-compatible widths & 9.42 \\
    Pruning form & Use body-only channel pruning without preserving attention-compatible widths & 41.85 \\
    Block-selection score & Replace the teacher-alignment score with an alternative pruning score & 8.16 \\
    Block selection & Select blocks greedily instead of using the proposed block set & 5.86 \\
    Block selection & Use an exhaustive candidate search under the same pruning family & 6.06 \\
    Repair noise & Use multi-noise block ranking and repair instead of fixed high-noise repair & 8.96 \\
    \bottomrule
  \end{tabular}
  \label{tab:ablations}
\end{table}

The ablations support the two main design choices. First, the repair bridge is not optional: direct SiDA from the pruned checkpoint fails badly, while teacher alignment gives the downstream distillation loss a usable starting point. Second, the pruning criterion should be tied to the teacher's local denoising behavior. Channel-only pruning, alternative scores, and multi-noise variants are all weaker than the proposed teacher-aware block selection followed by fixed high-noise repair.

\subsection{Benchmark Performance}

Table~\ref{tab:benchmark} compares class-conditional ImageNet-512 generation without classifier-free guidance. Most baseline values follow the SiDA comparison table~\cite{zhou2024sida}, with each method cited separately. We add the original EDM2-XS baseline and our repaired compact generators. Runtime is included as a hardware-specific reference: measured times use a single RTX 3090, and speedup is computed against EDM2-XS under the same measurement script.

\begin{table}[t]
  \caption{Benchmark performance on ImageNet-512 without classifier-free guidance. Dashes indicate values not reported in the source comparison or not measured under our setup.}
  \centering
  \resizebox{\textwidth}{!}{
  \begin{tabular}{lccccc}
    \toprule
    Method (CFG=N) & NFE ($\downarrow$) & \#Params & FID ($\downarrow$) & Inference Time ($\downarrow$) & Speedup \\
    \midrule
    ADM-G~\cite{dhariwal2021diffusion} & 250 & 559M & 23.24 & -- & -- \\
    U-DiT-B~\cite{tian2024udit} & 250 & 204M & 15.39 & -- & -- \\
    DiT-XL/2~\cite{peebles2022dit} & 250 & 675M & 12.03 & -- & -- \\
    MaskDiT~\cite{zheng2023maskdit} & 79 & 736M & 10.79 & -- & -- \\
    ADM-U~\cite{dhariwal2021diffusion} & 250 & \textgreater 559M & 9.96 & -- & -- \\
    LEGO-XL-PR~\cite{zheng2024learning} & 250 & 681M & 9.01 & -- & -- \\
    BigGAN-deep~\cite{brock2019biggan} & 1 & 160M & 8.43 & -- & -- \\
    DiMR-XL/3R~\cite{liu2024dimr} & 250 & 525M & 7.93 & -- & -- \\
    MaskGIT~\cite{chang2022maskgit} & 12 & 227M & 7.32 & -- & -- \\
    MAGVIT-v2~\cite{yu2023magvitv2} & 12 & 307M & 4.61 & -- & -- \\
    RIN~\cite{jabri2022rin} & 1000 & 320M & 3.95 & -- & -- \\
    MAGVIT-v2~\cite{yu2023magvitv2} & 64 & 307M & 3.07 & -- & -- \\
    VDM++~\cite{kingma2023vdmpp} & 512 & 2B & 2.99 & -- & -- \\
    MAR~\cite{li2024mar} & 100 & 481M & 2.74 & -- & -- \\
    StyleGAN-XL~\cite{sauer2022styleganxl} & $1{\times}2$ & 168M & 2.41 & -- & -- \\
    SiDA-EDM2-XS~\cite{zhou2024sida} & 1 & 125M & $2.228 \pm 0.037$ & -- & -- \\
    EDM2-XS~\cite{karras2024edm2} & 63 & 124.713M & 3.53 & 3390.8 ms/img & $1.0{\times}$ \\
    Ours (20\% pruned) & 1 & 98.826M & 3.12 & 62.7 ms/img & $54.1{\times}$ \\
    Ours (30\% pruned) & 1 & 88.029M & 4.26 & 51.6 ms/img & $65.7{\times}$ \\
    \bottomrule
  \end{tabular}
  }
  \label{tab:benchmark}
\end{table}

Qualitative samples from the repaired compact generators are shown in Appendix~\ref{app:qualitative}.

To test whether the repair bridge is tied to SiDA, we also apply it with Diff-Instruct~\cite{luo2023diffinstruct} on CIFAR-10. This experiment asks whether a different step-distillation objective can accept a pruned student after the same short teacher-alignment stage. Table~\ref{tab:cifar10-benchmark} reports the CIFAR-10 unconditional benchmark with the inception score column removed; the CIFAR-10 pruning and repair settings are given in Appendix~\ref{app:reproducibility}.

\begin{table}[t]
  \caption{Unconditional sample quality on CIFAR-10. Runtime and GPU memory are reported for measured rows with batch size 128. $^\ast$Methods that require synthetic data construction for distillation. $^\dagger$Methods that require real data for distillation.}
  \centering
  \resizebox{\textwidth}{!}{
  \begin{tabular}{lcccccc}
    \toprule
    Method & NFE ($\downarrow$) & \#Params & FID ($\downarrow$) & GPU Memory ($\downarrow$) & Inference Time ($\downarrow$) & Speedup \\
    \midrule
    \multicolumn{7}{l}{\textbf{Multiple Steps (include Diffusion Distillation)}} \\
    DDPM~\cite{ho2020ddpm} & 1000 & -- & 3.17 & -- & -- & -- \\
    LSGM~\cite{vahdat2021lsgm} & 147 & -- & 2.10 & -- & -- & -- \\
    PFGM~\cite{xu2022pfgm} & 110 & -- & 2.35 & -- & -- & -- \\
    EDM~\cite{karras2022edm} & 35 & 55.734M & 1.97 & 1584.1 MB & 36.953 s & $1.000{\times}$ \\
    DDIM~\cite{song2020ddim} & 50 & -- & 4.67 & -- & -- & -- \\
    DDIM~\cite{song2020ddim} & 10 & -- & 8.23 & -- & -- & -- \\
    DPM-Solver-2~\cite{lu2022dpmsolver} & 12 & -- & 5.28 & -- & -- & -- \\
    DPM-Solver-3~\cite{lu2022dpmsolver} & 12 & -- & 6.03 & -- & -- & -- \\
    3-DEIS~\cite{zhang2022deis} & 10 & -- & 4.17 & -- & -- & -- \\
    UniPC~\cite{zhao2023unipc} & 8 & -- & 5.10 & -- & -- & -- \\
    UniPC~\cite{zhao2023unipc} & 5 & -- & 23.22 & -- & -- & -- \\
    Denoise Diffusion GAN (T=2)~\cite{xiao2021denoisinggan} & 2 & -- & 4.08 & -- & -- & -- \\
    PD~\cite{salimans2022progressive} & 2 & -- & 5.58 & -- & -- & -- \\
    CT~\cite{song2023consistency} & 2 & -- & 5.83 & -- & -- & -- \\
    CD$^\dagger$~\cite{song2023consistency} & 2 & -- & 2.93 & -- & -- & -- \\
    \midrule
    \multicolumn{7}{l}{\textbf{Single Step}} \\
    Denoise Diffusion GAN (T=1)~\cite{xiao2021denoisinggan} & 1 & -- & 14.6 & -- & -- & -- \\
    KD$^\ast$~\cite{luhman2021knowledge} & 1 & -- & 9.36 & -- & -- & -- \\
    TDPM~\cite{zheng2022tdpm} & 1 & -- & 8.91 & -- & -- & -- \\
    1-ReFlow~\cite{liu2022rectifiedflow} & 1 & -- & 378 & -- & -- & -- \\
    CT~\cite{song2023consistency} & 1 & -- & 8.70 & -- & -- & -- \\
    1-ReFlow (+distill)$^\ast$~\cite{liu2022rectifiedflow} & 1 & -- & 6.18 & -- & -- & -- \\
    2-ReFlow (+distill)$^\ast$~\cite{liu2022rectifiedflow} & 1 & -- & 4.85 & -- & -- & -- \\
    3-ReFlow (+distill)$^\ast$~\cite{liu2022rectifiedflow} & 1 & -- & 5.21 & -- & -- & -- \\
    PD~\cite{salimans2022progressive} & 1 & -- & 8.34 & -- & -- & -- \\
    CD-L2$^\dagger$~\cite{song2023consistency} & 1 & -- & 7.90 & -- & -- & -- \\
    CD-LPIPS$^\dagger$~\cite{song2023consistency} & 1 & -- & 3.55 & -- & -- & -- \\
    Diff-Instruct~\cite{luo2023diffinstruct} & 1 & 55.73M & 4.53 & 1576.1 MB & 1.060 s & $34.853{\times}$ \\
    Ours (34\% pruned) & 1 & 36.807M & 5.32 & 1253.3 MB & 0.921 s & $40.118{\times}$ \\
    \bottomrule
  \end{tabular}
  }
  \label{tab:cifar10-benchmark}
\end{table}

\FloatBarrier

\section{Limitations}

The method still depends on the downstream distillation algorithm. We present repair as a general teacher-alignment step, but the current implementation mainly covers one-step latent generation with SiDA on EDM2-XS/ImageNet-512 and Diff-Instruct on CIFAR-10. Multi-step distillation remains open and may need a different repair noise schedule. The repair stage also introduces choices such as $\sigma_r$, calibration data, update count, and candidate block set; although it is much shorter than full retraining, it is still an added stage and currently needs real images or representative latents.

\section{Conclusion}

We reframed post-pruning recovery as a distillation problem. A direct replacement of retraining with one-step distillation leaves the pruned generator too misaligned with the teacher to produce usable samples. The proposed repair stage reduces this local mismatch before distillation, allowing the compact generator to train as a one-step model. On EDM2-XS/ImageNet-512, the repaired 20\% pruned generator improves FID over the original EDM2-XS baseline while reducing NFE from 63 to 1. These results suggest that post-pruning recovery can be redirected from many-step retraining to compact one-step generation.

\bibliographystyle{unsrt}
\bibliography{references}

@article{zhou2024sida,
  title={Adversarial Score Identity Distillation: Rapidly Surpassing the Teacher in One Step},
  author={Zhou, Mingyuan and Zheng, Huangjie and Gu, Yi and Wang, Zhendong and Huang, Hai},
  journal={arXiv preprint arXiv:2410.14919},
  year={2024}
}

@inproceedings{ho2020ddpm,
  title={Denoising Diffusion Probabilistic Models},
  author={Ho, Jonathan and Jain, Ajay and Abbeel, Pieter},
  booktitle={Advances in Neural Information Processing Systems},
  year={2020}
}

@inproceedings{dhariwal2021diffusion,
  title={Diffusion Models Beat GANs on Image Synthesis},
  author={Dhariwal, Prafulla and Nichol, Alex},
  booktitle={Advances in Neural Information Processing Systems},
  year={2021}
}

@inproceedings{rombach2022ldm,
  title={High-Resolution Image Synthesis with Latent Diffusion Models},
  author={Rombach, Robin and Blattmann, Andreas and Lorenz, Dominik and Esser, Patrick and Ommer, Bj{\"o}rn},
  booktitle={IEEE/CVF Conference on Computer Vision and Pattern Recognition},
  year={2022}
}

@article{karras2024edm2,
  title={Analyzing and Improving the Training Dynamics of Diffusion Models},
  author={Karras, Tero and Aittala, Miika and Lehtinen, Jaakko and Hellsten, Janne and Aila, Timo and Laine, Samuli},
  journal={arXiv preprint arXiv:2312.02696},
  year={2023}
}

@article{han2015learning,
  title={Learning both Weights and Connections for Efficient Neural Networks},
  author={Han, Song and Pool, Jeff and Tran, John and Dally, William J.},
  journal={arXiv preprint arXiv:1506.02626},
  year={2015}
}

@inproceedings{fang2023diffpruning,
  title={Structural Pruning for Diffusion Models},
  author={Fang, Gongfan and Ma, Xinyin and Wang, Xinchao},
  booktitle={Advances in Neural Information Processing Systems},
  year={2023}
}

@article{ramesh2025efficient,
  title={Efficient Pruning of Text-to-Image Models: Insights from Pruning Stable Diffusion},
  author={Ramesh, Samarth N. and Zhao, Zhixue},
  journal={arXiv preprint arXiv:2411.15113},
  year={2024}
}

@inproceedings{kim2023bksdm,
  title={{BK-SDM}: A Lightweight, Fast, and Cheap Version of Stable Diffusion},
  author={Kim, Bo-Kyeong and Song, Hyoung-Kyu and Castells, Thibault and Choi, Shinkook},
  booktitle={European Conference on Computer Vision},
  year={2024}
}

@inproceedings{li2023snapfusion,
  title={SnapFusion: Text-to-Image Diffusion Model on Mobile Devices within Two Seconds},
  author={Li, Yanyu and Wang, Huan and Jin, Qing and Hu, Ju and Chemerys, Pavlo and Fu, Yun and Wang, Yanzhi and Tulyakov, Sergey and Ren, Jian},
  booktitle={Advances in Neural Information Processing Systems},
  year={2023}
}

@inproceedings{salimans2022progressive,
  title={Progressive Distillation for Fast Sampling of Diffusion Models},
  author={Salimans, Tim and Ho, Jonathan},
  booktitle={International Conference on Learning Representations},
  year={2022}
}

@inproceedings{song2023consistency,
  title={Consistency Models},
  author={Song, Yang and Dhariwal, Prafulla and Chen, Mark and Sutskever, Ilya},
  booktitle={International Conference on Machine Learning},
  year={2023}
}

@inproceedings{yin2024dmd,
  title={One-step Diffusion with Distribution Matching Distillation},
  author={Yin, Tianwei and Gharbi, Micha{\"e}l and Zhang, Richard and Shechtman, Eli and Durand, Fr{\'e}do and Freeman, William T. and Park, Taesung},
  booktitle={IEEE/CVF Conference on Computer Vision and Pattern Recognition},
  year={2024}
}

@article{castells2024ldpruner,
  title={{LD-Pruner}: Efficient Pruning of Latent Diffusion Models using Task-Agnostic Insights},
  author={Castells, Thibault and Song, Hyoung-Kyu and Kim, Bo-Kyeong and Choi, Shinkook},
  journal={arXiv preprint arXiv:2404.11936},
  year={2024}
}

@article{zhang2024effortless,
  title={Learnable Sparsity for Vision Generative Models},
  author={Zhang, Yang and Jin, Er and Liang, Wenzhong and Dong, Yanfei and Khakzar, Ashkan and Torr, Philip and Stegmaier, Johannes and Kawaguchi, Kenji},
  journal={arXiv preprint arXiv:2412.02852},
  year={2024}
}

@inproceedings{tian2024udit,
  title={{U-DiTs}: Downsample Tokens in U-Shaped Diffusion Transformers},
  author={Tian, Yuchuan and Tu, Zhijun and Chen, Hanting and Hu, Jie and Xu, Chao and Wang, Yunhe},
  booktitle={Advances in Neural Information Processing Systems},
  year={2024}
}

@inproceedings{peebles2022dit,
  title={Scalable Diffusion Models with Transformers},
  author={Peebles, William and Xie, Saining},
  booktitle={International Conference on Computer Vision},
  year={2023}
}

@article{zheng2023maskdit,
  title={Fast Training of Diffusion Models with Masked Transformers},
  author={Zheng, Hongkai and Nie, Weili and Vahdat, Arash and Anandkumar, Anima},
  journal={Transactions on Machine Learning Research},
  year={2024}
}

@article{zheng2024learning,
  title={Learning Stackable and Skippable {LEGO} Bricks for Efficient, Reconfigurable, and Variable-Resolution Diffusion Modeling},
  author={Zheng, Huangjie and Wang, Zhendong and Yuan, Jianbo and Ning, Guanghan and He, Pengcheng and You, Quanzeng and Yang, Hongxia and Zhou, Mingyuan},
  journal={International Conference on Learning Representations},
  year={2024}
}

@inproceedings{brock2019biggan,
  title={Large Scale {GAN} Training for High Fidelity Natural Image Synthesis},
  author={Brock, Andrew and Donahue, Jeff and Simonyan, Karen},
  booktitle={International Conference on Learning Representations},
  year={2019}
}

@article{liu2024dimr,
  title={Alleviating Distortion in Image Generation via Multi-Resolution Diffusion Models and Time-Dependent Layer Normalization},
  author={Liu, Qihao and Zeng, Zhanpeng and He, Ju and Yu, Qihang and Shen, Xiaohui and Chen, Liang-Chieh},
  journal={arXiv preprint arXiv:2406.09416},
  year={2024}
}

@inproceedings{chang2022maskgit,
  title={{MaskGIT}: Masked Generative Image Transformer},
  author={Chang, Huiwen and Zhang, Han and Jiang, Lu and Liu, Ce and Freeman, William T.},
  booktitle={IEEE/CVF Conference on Computer Vision and Pattern Recognition},
  year={2022}
}

@inproceedings{yu2023magvitv2,
  title={Language Model Beats Diffusion -- Tokenizer is Key to Visual Generation},
  author={Yu, Lijun and Lezama, Jos{\'e} and Gundavarapu, Nitesh B. and Versari, Luca and Sohn, Kihyuk and Minnen, David and Cheng, Yong and Birodkar, Vighnesh and Gupta, Agrim and Gu, Xiuye and Hauptmann, Alexander G. and Gong, Boqing and Yang, Ming-Hsuan and Essa, Irfan and Ross, David A. and Jiang, Lu},
  booktitle={International Conference on Learning Representations},
  year={2024}
}

@inproceedings{jabri2022rin,
  title={Scalable Adaptive Computation for Iterative Generation},
  author={Jabri, Allan and Fleet, David and Chen, Ting},
  booktitle={International Conference on Machine Learning},
  year={2023}
}

@article{kingma2023vdmpp,
  title={Understanding Diffusion Objectives as the {ELBO} with Simple Data Augmentation},
  author={Kingma, Diederik P. and Gao, Ruiqi},
  journal={arXiv preprint arXiv:2303.00848},
  year={2023}
}

@inproceedings{li2024mar,
  title={Autoregressive Image Generation without Vector Quantization},
  author={Li, Tianhong and Tian, Yonglong and Li, He and Deng, Mingyang and He, Kaiming},
  booktitle={Advances in Neural Information Processing Systems},
  year={2024}
}

@article{sauer2022styleganxl,
  title={{StyleGAN-XL}: Scaling {StyleGAN} to Large Diverse Datasets},
  author={Sauer, Axel and Schwarz, Katja and Geiger, Andreas},
  journal={ACM Transactions on Graphics},
  year={2022}
}

@inproceedings{vahdat2021lsgm,
  title={Score-Based Generative Modeling in Latent Space},
  author={Vahdat, Arash and Kreis, Karsten and Kautz, Jan},
  booktitle={Advances in Neural Information Processing Systems},
  year={2021}
}

@inproceedings{xu2022pfgm,
  title={Poisson Flow Generative Models},
  author={Xu, Yilun and Liu, Ziming and Tegmark, Max and Jaakkola, Tommi},
  booktitle={Advances in Neural Information Processing Systems},
  year={2022}
}

@inproceedings{karras2022edm,
  title={Elucidating the Design Space of Diffusion-Based Generative Models},
  author={Karras, Tero and Aittala, Miika and Aila, Timo and Laine, Samuli},
  booktitle={Advances in Neural Information Processing Systems},
  year={2022}
}

@inproceedings{song2020ddim,
  title={Denoising Diffusion Implicit Models},
  author={Song, Jiaming and Meng, Chenlin and Ermon, Stefano},
  booktitle={International Conference on Learning Representations},
  year={2021}
}

@inproceedings{lu2022dpmsolver,
  title={{DPM-Solver}: A Fast {ODE} Solver for Diffusion Probabilistic Model Sampling in Around 10 Steps},
  author={Lu, Cheng and Zhou, Yuhao and Bao, Fan and Chen, Jianfei and Li, Chongxuan and Zhu, Jun},
  booktitle={Advances in Neural Information Processing Systems},
  year={2022}
}

@inproceedings{zhang2022deis,
  title={Fast Sampling of Diffusion Models with Exponential Integrator},
  author={Zhang, Qinsheng and Chen, Yongxin},
  booktitle={International Conference on Learning Representations},
  year={2023}
}

@inproceedings{zhao2023unipc,
  title={{UniPC}: A Unified Predictor-Corrector Framework for Fast Sampling of Diffusion Models},
  author={Zhao, Wenliang and Bai, Lujia and Rao, Yongming and Zhou, Jie and Lu, Jiwen},
  booktitle={Advances in Neural Information Processing Systems},
  year={2023}
}

@inproceedings{xiao2021denoisinggan,
  title={Tackling the Generative Learning Trilemma with Denoising Diffusion {GANs}},
  author={Xiao, Zhisheng and Kreis, Karsten and Vahdat, Arash},
  booktitle={International Conference on Learning Representations},
  year={2022}
}

@inproceedings{luhman2021knowledge,
  title={Knowledge Distillation in Iterative Generative Models for Improved Sampling Speed},
  author={Luhman, Eric and Luhman, Troy},
  booktitle={arXiv preprint arXiv:2101.02388},
  year={2021}
}

@inproceedings{zheng2022tdpm,
  title={Truncated Diffusion Probabilistic Models and Diffusion-based Adversarial Auto-Encoders},
  author={Zheng, Huangjie and Nie, Weili and Vahdat, Arash and Azizzadenesheli, Kamyar and Anandkumar, Anima},
  booktitle={International Conference on Learning Representations},
  year={2022}
}

@article{liu2022rectifiedflow,
  title={Flow Straight and Fast: Learning to Generate and Transfer Data with Rectified Flow},
  author={Liu, Xingchao and Gong, Chengyue and Liu, Qiang},
  journal={arXiv preprint arXiv:2209.03003},
  year={2022}
}

@article{luo2023diffinstruct,
  title={Diff-Instruct: A Universal Approach for Transferring Knowledge From Pre-trained Diffusion Models},
  author={Luo, Weijian and Hu, Tianyang and Zhang, Shifeng and Sun, Jiacheng and Li, Zhenguo and Zhang, Zhihua},
  journal={arXiv preprint arXiv:2305.18455},
  year={2023}
}

\clearpage

\appendix
\section*{Appendix}

\section{Reproducibility Details}
\label{app:reproducibility}

The main text keeps the experimental narrative at the protocol level. This appendix records the run-level settings needed to reproduce the reported pruning and repair runs.

\subsection{ImageNet-512 Settings}

For ImageNet-512, the teacher is EDM2-XS with 124.713M parameters. The two compact generators reported in Table~\ref{tab:benchmark} have 98.826M and 88.029M parameters. Block sensitivity is estimated at $\sigma^\star=2.5$ with 8 calibration batches and 8 samples per batch. The teacher-alignment repair stage uses $\sigma_r=2.5$ and an effective batch size of 128. After repair, SiDA is run with the frozen EDM2 teacher as the true score model and a fake-score network initialized from the teacher.

For exact replication of the more compressed ImageNet model in Table~\ref{tab:benchmark}, the disabled EDM2 U-Net branches are listed below. These names refer to implementation modules in the EDM2 U-Net and are not part of the method definition.

\begin{table}[H]
  \caption{Disabled branches for the more compressed ImageNet-512 compact generator.}
  \centering
  \small
  \begin{tabular}{ll}
    \toprule
    U-Net side & Disabled branches \\
    \midrule
    Encoder & \texttt{enc.8x8\_block0}, \texttt{enc.8x8\_block1}, \texttt{enc.8x8\_block2} \\
    Decoder & \texttt{dec.8x8\_in0}, \texttt{dec.8x8\_block3}, \texttt{dec.16x16\_block1} \\
    \bottomrule
  \end{tabular}
  \label{tab:imagenet-disabled-branches}
\end{table}

\subsection{CIFAR-10 Diff-Instruct Settings}

For the CIFAR-10 experiment in Table~\ref{tab:cifar10-benchmark}, we start from the public EDM CIFAR-10 unconditional VP checkpoint and use Diff-Instruct as the downstream one-step distillation objective. The dataset is CIFAR-10 at $32{\times}32$ resolution, with 50k training images, no labels, and no horizontal flipping. The teacher architecture is EDMPrecond SongUNet with 128 base channels, channel multipliers $(2,2,2)$, dropout 0.13, and $\sigma_{\mathrm{init}}=1.0$.

\begin{table}[H]
  \caption{Common CIFAR-10 distillation settings used with Diff-Instruct.}
  \centering
  \small
  \begin{tabular}{ll}
    \toprule
    Item & Setting \\
    \midrule
    Teacher checkpoint & EDM CIFAR-10 unconditional VP checkpoint \\
    Distillation loss & Diff-Instruct \texttt{DI\_EDMLoss} \\
    Generator optimizer & Adam, learning rate $1{\times}10^{-5}$, betas $(0,0.999)$, $\epsilon=10^{-8}$ \\
    Score-generator optimizer & Adam, learning rate $1{\times}10^{-5}$, betas $(0,0.999)$, $\epsilon=10^{-8}$ \\
    Training batch size & 128 \\
    EMA half-life & 500 kimg \\
    Evaluation metric & FID-50k full \\
    \bottomrule
  \end{tabular}
  \label{tab:cifar-common-settings}
\end{table}

\begin{table}[H]
  \caption{CIFAR-10 pruning and repair settings for the compact Diff-Instruct students.}
  \centering
  \small
  \resizebox{\textwidth}{!}{
  \begin{tabular}{lll}
    \toprule
    Item & 18\% compact student & 34\% compact student \\
    \midrule
    Pruning scope and type & Generator-only width pruning & Generator-only width pruning \\
    Channel selection & Prefix EDM parameter Taylor score & Prefix EDM parameter Taylor score \\
    Target pruning ratio & 0.20 & 0.32 \\
    Selected base channels & 116 & 104 \\
    Parameter count & 45.78M from a 55.73M teacher & 36.81M from a 55.73M teacher \\
    Actual parameter reduction & 17.86\% & 33.96\% \\
    Protected channel fraction & 0.9 & 0.9 \\
    Calibration & 8 pruning batches; score batch 8; 2 Diff-Taylor batches & Same \\
    Diff-Taylor score & 16 steps with threshold 0.05 & Same \\
    Noise bins for pruning checks & $\{1.0, 0.3, 0.09\}$ & Same \\
    Repair schedule & 12k steps, batch 16, learning rate $2{\times}10^{-4}$ & Same \\
    Repair noise & Lognormal noise; max-loss selection over 16 sigma steps & Same \\
    Repair weights & Init/noisy weights $0/1$; score gen/mean/max weights $1/5/1$ & Same \\
    \bottomrule
  \end{tabular}
  }
  \label{tab:cifar-pruning-repair-settings}
\end{table}

\section{Additional Qualitative Results}
\label{app:qualitative}

\begin{figure}[H]
  \centering
  \includegraphics[width=0.72\textwidth]{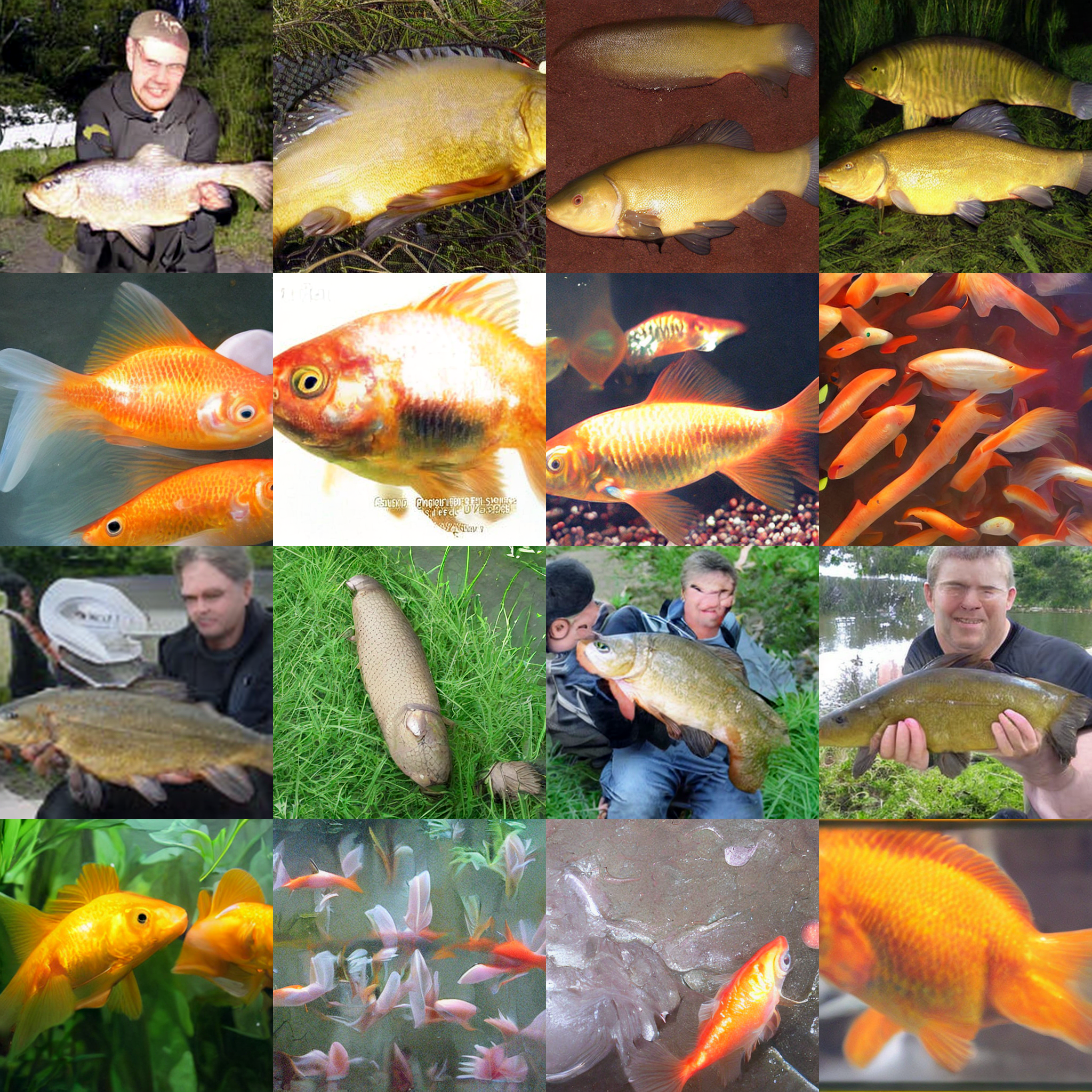}
  \caption{ImageNet-512 samples from our 20\% pruned one-step generator. This model reaches an FID of 3.12 with 98.826M parameters and one network evaluation.}
  \label{fig:imagenet-samples}
\end{figure}

\begin{figure}[H]
  \centering
  \includegraphics[width=0.72\textwidth]{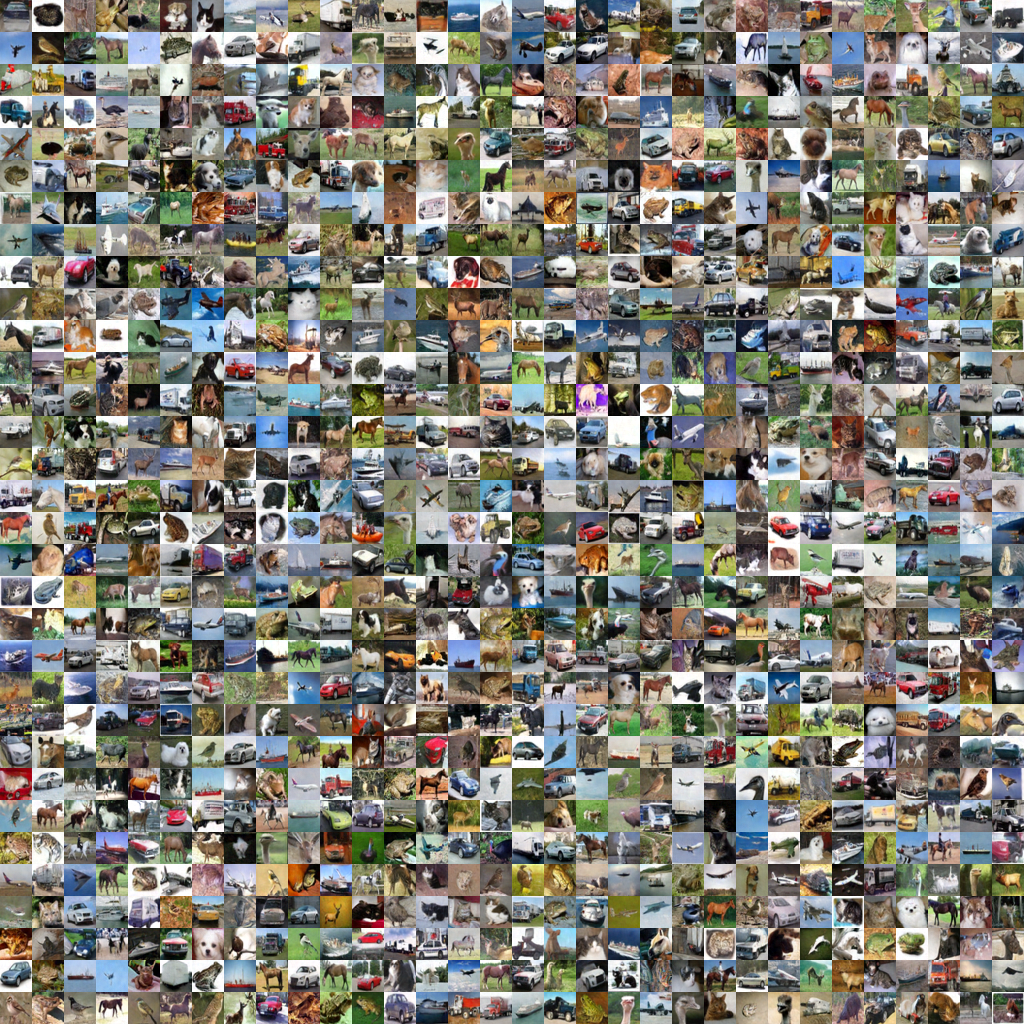}
  \caption{CIFAR-10 samples from our 34\% pruned one-step generator distilled with Diff-Instruct.}
  \label{fig:cifar10-samples}
\end{figure}

\end{document}